\documentclass[sigconf]{acmart}

\usepackage{booktabs}
\usepackage{graphicx}
\usepackage{amsmath,amssymb} 
\usepackage{bm}
\usepackage{subfigure}
\usepackage{multirow}
\usepackage{epstopdf}
\usepackage{url}
\usepackage{ulem}
\usepackage{color}
\usepackage{balance}

\copyrightyear{2017} 
\acmYear{2017} 
\setcopyright{acmcopyright}
\acmConference{MM '17}{October 23--27, 2017}{Mountain View, CA, USA}\acmPrice{15.00}\acmDOI{10.1145/3123266.3123431}
\acmISBN{978-1-4503-4906-2/17/10}

\begin{document}
\title{Face Aging with Contextual Generative Adversarial Nets} 

\author{Si Liu}
\affiliation{\institution{SKLOIS, IIE, CAS}}
\email{liusi@iie.ac.cn}

\author{Yao Sun}
\affiliation{\institution{SKLOIS, IIE, CAS}}
\authornote{corresponding author}
\email{sunyao@iie.ac.cn}

\author{Defa Zhu}
\affiliation{\institution{SKLOIS, IIE, CAS}}
\affiliation{\institution{School of Cyber Security, UCAS}}
\email{18502408950@163.com}

\author{Renda Bao}
\affiliation{\institution{SKLOIS, IIE, CAS}}
\affiliation{\institution{School of Cyber Security, UCAS}}
\email{roger\_bao@163.com}

\author{Wei Wang}
\affiliation{\institution{University of Trento, Italy}}
\email{wangwei1990@gmail.com}

\author{Xiangbo Shu}
\affiliation{\institution{Nanjing University of Science and Technology}}
\email{shuxb@njust.edu.cn}

\author{Shuicheng Yan}
\affiliation{\institution{Qihoo 360 AI Institute, Beijing, China}}
\affiliation{\institution{National University of singapore}}
\email{eleyans@nus.edu.sg}

	\begin{abstract}
Face aging, which renders aging faces for an input face, has attracted extensive attention in the multimedia research. Recently, several conditional Generative Adversarial Nets (GANs) based methods have achieved great success. They can generate images fitting the real face distributions conditioned on each individual age group. However, these methods fail to capture the transition patterns, e.g., the gradual shape and texture changes between adjacent age groups. In this paper, we propose a novel Contextual Generative Adversarial Nets (C-GANs) to specifically take it into consideration. The C-GANs consists of a conditional transformation network and two discriminative networks. The conditional transformation network imitates the aging procedure with several specially designed residual blocks. The age discriminative network guides the synthesized face to fit the real conditional distribution. The transition pattern discriminative network is novel, aiming to distinguish the real transition patterns with the fake ones. It serves as an extra regularization term for the conditional transformation network, ensuring the generated image pairs to fit the corresponding real transition pattern distribution. Experimental results demonstrate the proposed framework produces appealing results by comparing with the state-of-the-art and ground truth. We also observe performance gain for cross-age face verification.		
	\end{abstract}

	%
	%
	
\keywords{Face Aging, Generative Adversarial Nets, 	Contextual Modeling}

	\maketitle

\begin{figure}[t]
	
	\begin{center}
		\includegraphics[width=1\linewidth]{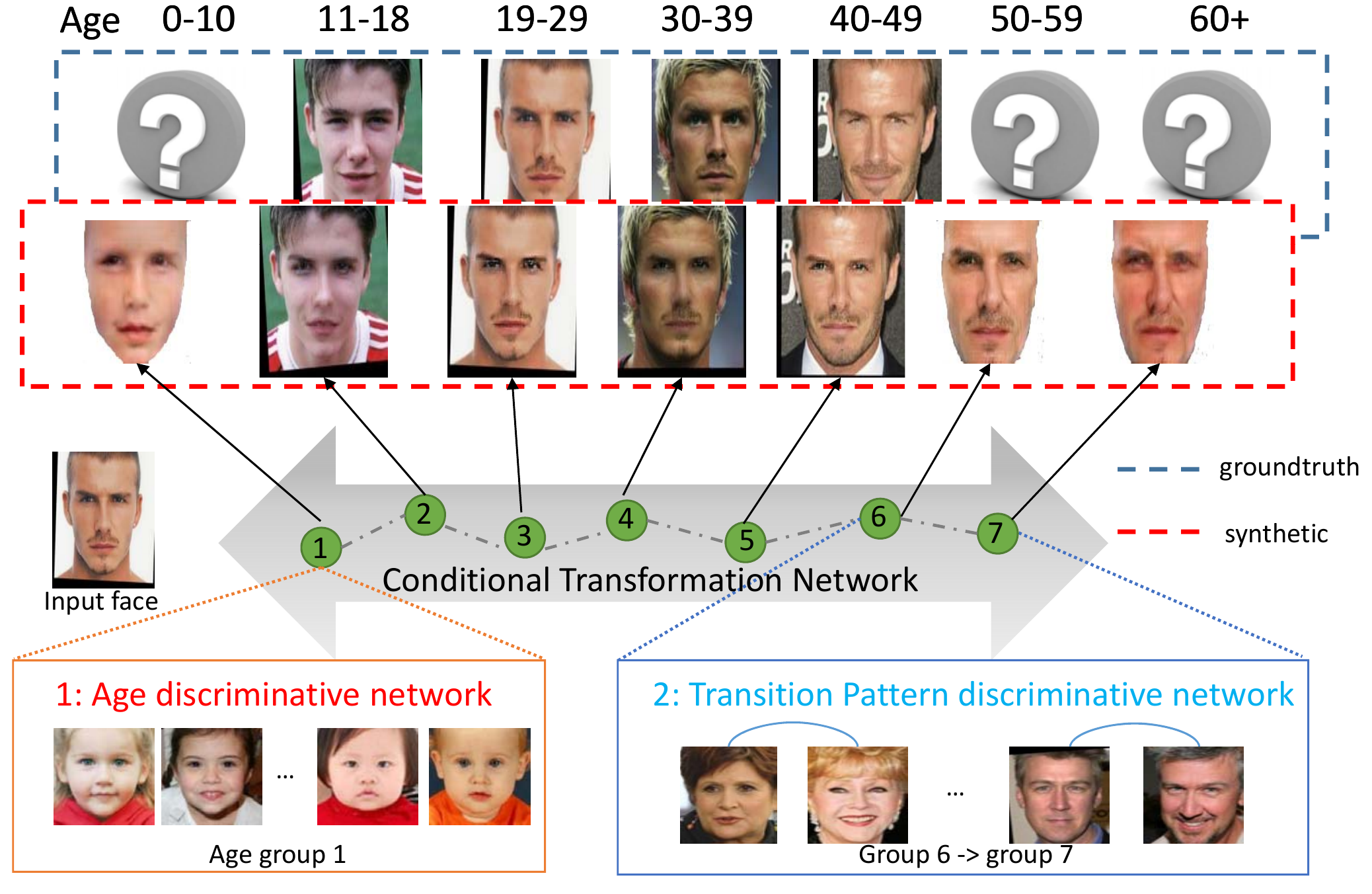}
		\caption{The proposed C-GANs algorithm for face aging. The input image can be  transformed to any specific age group. The synthesized results of C-GANs are natural due to the two discriminative networks:  the age discriminative network modeling  the distribution of each individual age group, while the transition pattern discriminative network modeling the correlations between adjacent groups.   }
		\label{fig:firstfig}
	\end{center}
  \vspace{-4mm}
\end{figure}

\begin{figure*}[t]
	\begin{center}
		\includegraphics[width=1\linewidth]{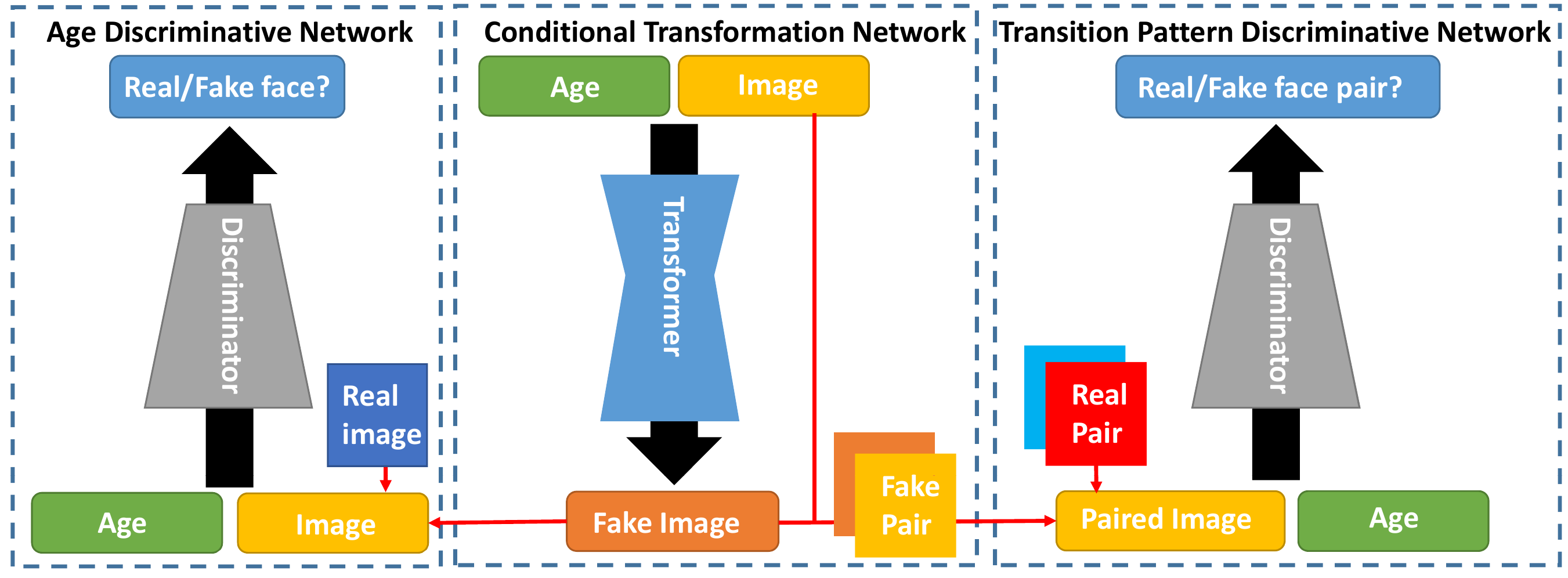}
		\caption{The structure of the proposed C-GANs.}
		\label{fig:framework}
	\end{center}
  \vspace{-4mm}
\end{figure*}

\section{Introduction}
Face aging, also known as age progression \cite{Shu2015Personalized,demo}, is attracting more and more research interests. It has plenty of applications  in various domains including cross-age face recognition \cite{Park2010Age}, finding lost children,  and entertainments \cite{Wang2016Recurrent}. In recent years, face aging has witnessed various breakthroughs and a number of face aging models have been proposed \cite{Fu2010Age}. Face aging, however, is still a very challenging task in practice for various reasons. First, faces may have many different expressions and lighting conditions, which pose great challenges to modeling the aging patterns. Besides, the training data are usually very limited and the face images for the same person only cover a narrow range of ages.

Traditional face aging approaches can be roughly split into to two classes, i.e., the prototyping ones \cite{Kemelmacher2014Illumination,Tiddeman2001Prototyping},  and the modeling ones \cite{Suo2010A,Tazoe2012Facial}. However, these approaches often require face aging sequences of the same person with wide range of ages which are very costly to collect. Generative Adversarial Networks (GANs) \cite{Goodfellow2014Generative}  better deal with age progressions. Many GANs based methods  \cite{Antipov2017Face,Zhang2017Age} can generate the most plausible and realistic images which are hard to distinguish from real data conditioned on the age. However, all of these methods do not make full use of the sequential data. Therefore, these methods cannot explicitly consider the transition patterns which are defined as the facial feature correlations between different age groups for one person. Therefore, their results are usually unable to maintain face identity, or cannot satisfy the cross-aging transition rules well.

In this paper, we mainly consider the cross-age transition pattern. Specifically, transition pattern contains two aspects. One is the identity consistency, and the other is the  appearance changes. Identity preserving is critical in face aging based   applications, e.g., cross-age face verification. Appearance changes include texture and shape alterations. Transition pattern  is age-aware. For example, when one grows from baby to teenagers, the main appearance difference is the face becomes larger. When one grows from the age of $50$ to $60$, the main facial changes lie on the texture alteration, such as the gradually developed  eye bag, senile plaques and wrinkle. Different from traditional GANs which only model the real data distribution of each individual age,  we focus on the higher-order cross-age correlations, which will make the face aging results more appealing. To model the above-mentioned transition  patterns, we propose a Contextual Generative Adversarial Nets (C-GANs). Figure \ref{fig:firstfig} illustrates C-GANs briefly. For an input face, C-GANs can generate faces for any target age group. To ensure the generated images real, C-GANs uses two discriminative networks to model the distribution of each individual age group as well as the transition patterns of two adjacent groups respectively.

More specifically, C-GANs consists of three networks, which is shown in Figure \ref{fig:framework}. The conditional transformation network transforms the input face to the desired age; the age discriminative network assists generating images indistinguishable with the real ones; the transition pattern discriminative network regularize the generated images satisfying the cross-age aging rules. The proposed C-GANs can be trained end-to-end and very easy to reproduce. In order to facilitate the presentation, we only mention  face aging/progression in this paper. Actually,  C-GANs can also achieve  face regression without any further modification.

\begin{figure*}[t!]
	\begin{center}
		\includegraphics[width=1\linewidth]{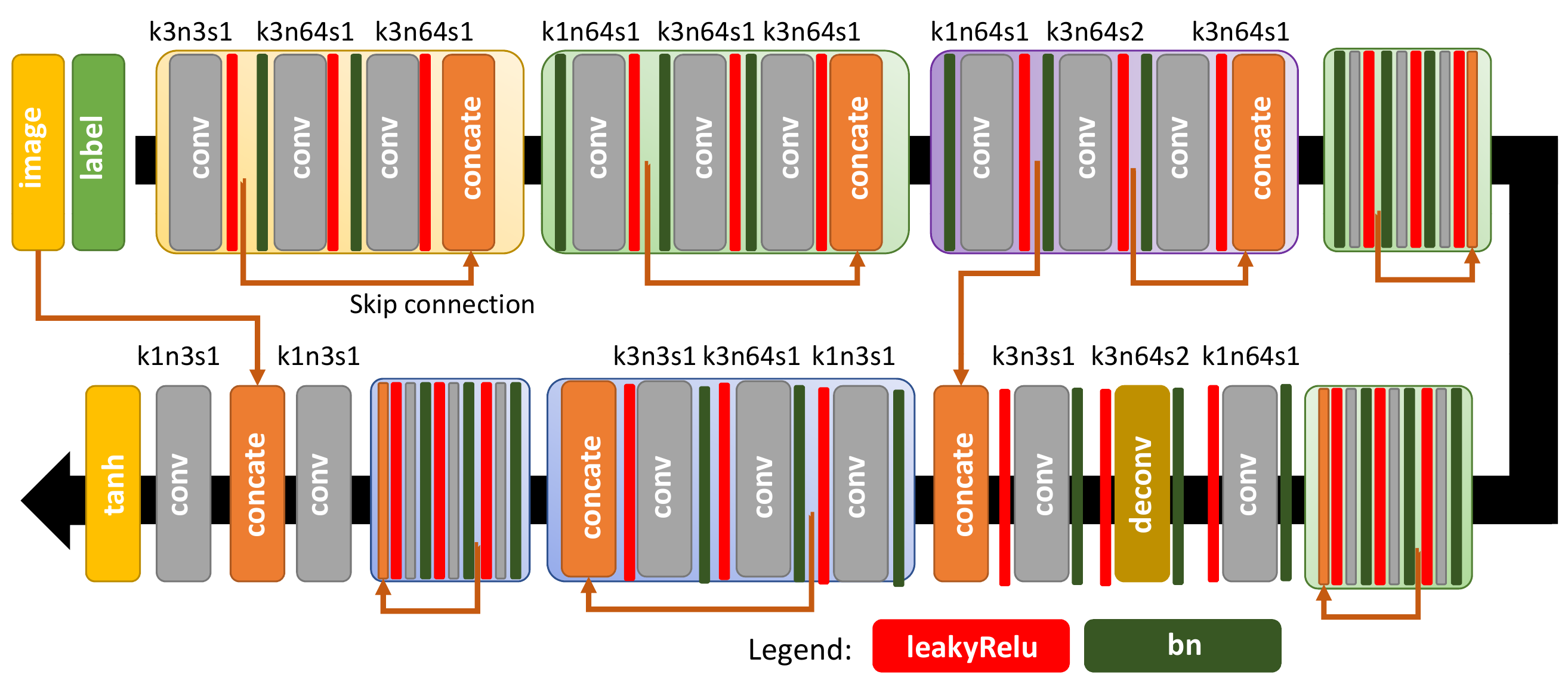}
		\vspace{-6mm}
		\caption{Architecture  of the conditional transformation network  $i$-th corresponding kernel size (k), number of feature maps 	(n) and stride (s) indicated for each convolutional layer.
		}
		\label{fig:generator}
	\end{center}
  \vspace{-4mm}
\end{figure*}

The contributions of this paper are summarized as follows. 
\begin{enumerate}
	\item We design an effective and efficient contextual GANs based face aging system whose aging results are significantly better than  existing face aging methods. The source code of our method will be released to the academic area for further research. 
	\item We introduce a novel transition pattern discriminative network in the C-GANs to regularize the synthesized faces satisfying the cross-age face aging rules.
	\item The conditional face transformation network in C-GANs is different with existing GANs generators in that it is much deeper, with several specially designed skip layers to preserve both the high-level semantics and low-level information. It makes the generated images more natural and real.
	
\end{enumerate}

\section{Related Works}

\subsection{Face Aging}

Traditional face aging models can be roughly divided into physical model approaches and prototype approaches.  Physical model approaches explicitly or implicitly models shape and texture  parameters for each age group. For example, Suo \emph{et al.} \cite{Suo2010A}  present a hierarchical And-Or graph based  dynamic model for face aging.  Other model-based age progression approaches include active appearance model \cite{Lanitis2002Toward}, support vector regression \cite{Patterson2007Aspects} and implicit function \cite{Berg2006A}. 
The prototype approache \cite{Kemelmacher2014Illumination} aim at constructing a relighted average face as prototypes for different age groups, and transferring the texture difference between the prototypes to the test image.  However, the limitation of this model is they are based on general rules, they totally discard the personalized information.   Recently,
Shu \emph{et al.} propose a coupled dictionary learning (CDL) model \cite{Shu2015Personalized}.  It encodes the aging patterns by the dictionary bases. Every two neighboring dictionaries are learned jointly.  However,
this method still has ghost artifacts as the reconstruction
residual does not evolve over time. Wang \emph{et al.}  \cite{Wang2016Recurrent}   introduce a recurrent face aging (RFA) framework based on a recurrent neural network. They employ a two-layer gated recurrent unit as the basic recurrent module whose bottom layer encodes a young face to a latent representation and the top layer decodes the representation to a corresponding older face. 
Generally, these techniques requires sufficient age sequences  as the training data, which limits these methods's practicality.

\subsection{Generative Adversarial Networks}

Recently, GANs \cite{Goodfellow2014Generative} has achieved great success in many image synthesis applications, including super resolution  \cite{Ledig2016Photo},  image-to-image translation by pix2pix  \cite{Isola2016Image} and   CycleGAN \cite{Zhu2017Unpaired}, in-paining \cite{Pathak2016Context},   visual manipulation on the images \cite{Zhu2016Generative}.

Antipo \emph{et al.} \cite{Antipov2017Face}  propose the GAN-based method for automatic face aging. They particularly emphasize  preserving the original person's identity by introducing a ``Identity-Preserving'' optimization of GAN's latent vectors.  Zhang  \cite{Zhang2017Age} propose a conditional adversarial autoencoder (CAAE) that learns a face manifold, traversing on which smooth age progression and regression can be realized simultaneously. Two adversarial networks are imposed on the encoder and generator, respectively, forcing to generate more photo-realistic faces. Li \emph{et al.} \cite{Li2016Deep} presents a Deep convolutional network model for Identity-Aware Transfer (DIAT) of facial attributes.  However, these GANs based methods  independently model the distribution of  each age group, without capturing the cross-age transition patterns.

\section{Approach}

The architecture of the proposed C-GANs is shown in Figure \ref{fig:framework}.
The input image  $x$ is first aligned and parsed  (Section \ref{sec:pre_processing}). 
Then $x$ is paired with an arbitrary  age label $\tilde y$ to feed into the conditional transformation network $G$ 
(Section \ref{sec:transformation_network}). 
The synthesized face ${G\left( {x,\tilde y} \right)}$  is judged by the age discriminative network $D_a$ to be real/fake (Section \ref{sec:discrimaintive_network}). Moreover, the age pair 
composed of a real image and its fake counterpart,  is fed into the transition pattern discriminative   network $D_t$ which predicts whether  it is from the  real image pair distribution (Section \ref{sec:transition_pattern}). 
Finally, the objective function and the training strategy is introduced (Section \ref{sec:object_function}).

\subsection{Image Preprocessing} \label{sec:pre_processing}

The input image $x$ is aligned via the face alignment techniques  which locates $68$ points on the faces. The landmark is used to align the faces. Then we use Deeplab v2 \cite{Chen2016DeepLab} to parse the  human face into  facial and non-facial regions. The non-facial region containing the background, the hair and clothes, are masked with gray color to facilitate the GANs training.

\subsection{Conditional Transformation Network}   \label{sec:transformation_network}

Given the input face $x$ and the desired age $\tilde y$, the Conditional Transformation Network generates the synthesized face  ${x_{\tilde y}} = G\left( {x,\tilde y} \right)$. The architecture of the generator is shown in  Figure \ref{fig:generator}. The input and output face images are $128$ $\times$ $128$ RGB images. 
The output is  in range $[-1,1]$     through the hyperbolic tangent function. Normalizing the input may make the training process converge faster. The conditions of C-GANs are 7-dim one-hot age vectors, and reshaped as 7-channel tensor with the same spatial dimensions with the input face.
The input faces and the labels are concatenated and fed into the further processing. To make fair concatenation, the elements of label is also confined to $[-1,1]$, where $-1$ corresponds to $0$.

The conditional transformation network mainly contains several residual blocks \cite{He2015Deep} with several skip layers.
Following DCGAN \cite{Radford2015Unsupervised},  the convolution of stride 2 is employed instead of pooling. 
The first three residual blocks donwsize the feature maps to half the resolution of the input image. 
The ``deconv''  layer upsamples the feature map to the original resolution. Such layers have been used previously
\cite{Zeiler2014Visualizing,Shelhamer2017Fully}. 
Note that to perform the refinement, we adopt several skip layer. 
In this way we preserve both the high-level information passed from coarser feature maps and fine local information provided in lower layer feature maps. 

\subsection{Age Discriminative Network}  \label{sec:discrimaintive_network}

\begin{figure}[t]
	\begin{center}
		\includegraphics[width=0.9\linewidth]{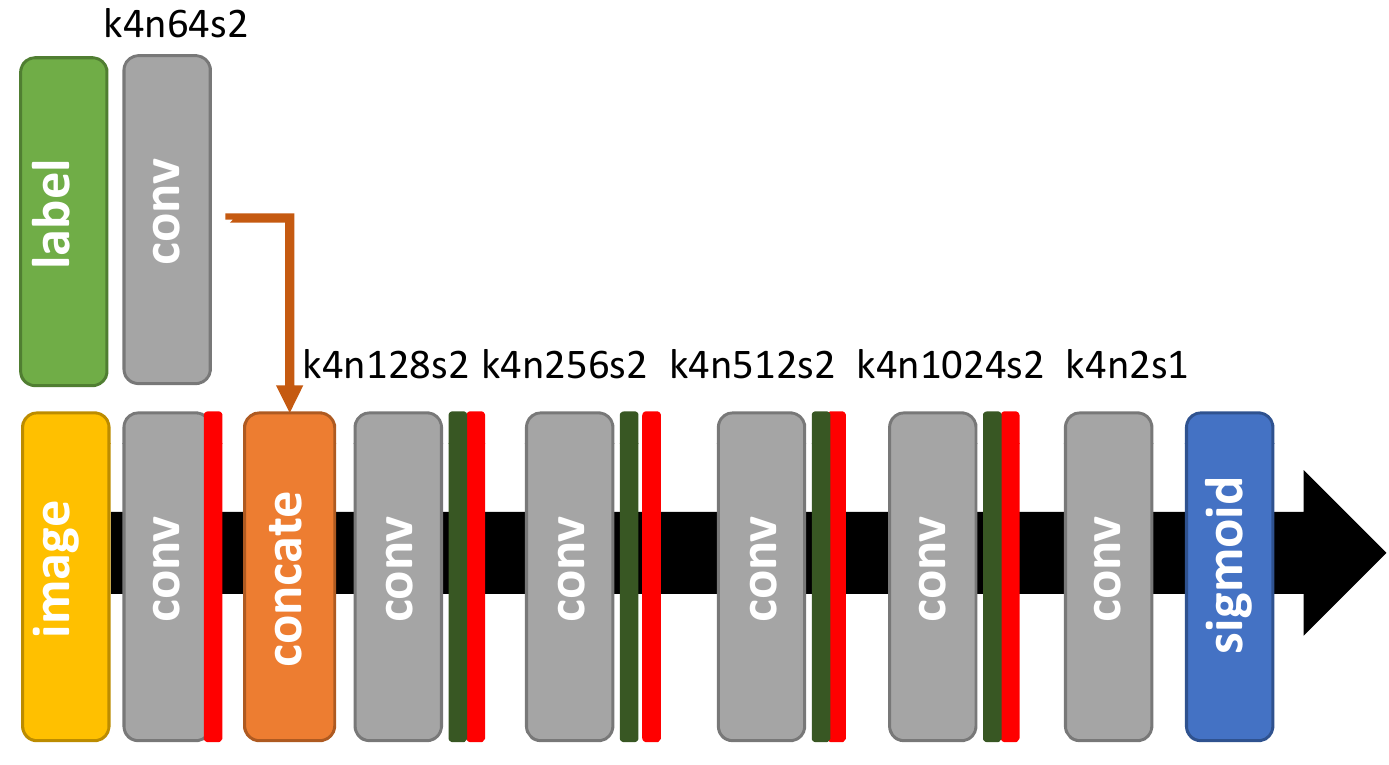}
		\vspace{-4mm}
		\caption{The structure of the age discriminative network.}
		\label{fig:D}
	\end{center}
  \vspace{-4mm}
\end{figure}

The  structure of the age discriminative network  
${{D_a}\left( {x,y} \right)}$ is shown in Figure \ref{fig:D}, similar to the conditional GAN \cite{Mirza2014Conditional}. 
${{\theta _{{D_a}}}}$ is the network parameters.
More formally, the training can be expressed as an optimization of the function $E({\theta _G},{{\theta _{{D_a}}}})$, where ${\theta _G}$ and ${{\theta _{{D_a}}}}$ are parameters of $G$ and $D_a$, respectively:

\begin{equation}
\begin{array}{*{20}{l}}
{\mathop {\min }\limits_G \mathop {\max }\limits_{{D_a}} E\left( {{\theta _G},{\theta _{{D_a}}}} \right) = {E_{{x_y},y\sim{p_{data}}({x_y},y)}}\left[ {\log {D_a}\left( {{x_y},y} \right)} \right]}\\
{ + {E_{x\sim{p_x},\tilde y\sim{p_y}}}\left[ {\log \left( {1 - {D_a}\left( {G\left( {x,\tilde y} \right),\tilde y} \right)} \right)} \right]}.
\end{array}
\end{equation}

Note that the age label is resized to a tensor similar with the conditional transformation network. 
The image and the labels go through one convolution layers individually  and concatenated to feed to  ${D_a}$ to make it discriminative on both age and human face.  During training, 
the positive samples are the real faces and their corresponding age $\left\{ {x_y,y} \right\}$, while nagetive samples  are $\{ {x_{\tilde y}},{\tilde y}\}$, where ${x_{\tilde y}}$ is the generated/fake face  and  ${\tilde y}$ is the corrsponding label used during the generation process. Note that we specifically randomly sample the label ${\tilde y}$ for the fake images to enhance the generalization ability of the C-GANs model.

\subsection{Transition Pattern Discriminative Network}  \label{sec:transition_pattern}

For better face aging results, we specifically model the cross-age transition patterns, which is defined as the facial feature correlations betweeen different age groups. 
In this paper, we only  consider adjacent age groups for simplicity. The long-range transition pattern can be  represented as the combination of a series of transitions patterns between adjacent age group.  
The transion pattern is age-aware. As shown in Figure \ref{fig:D},  when a person grows from the age of $10$  to $20$, the facial shape  alters as the skull grows. However, when one grows from the age of $50$ to $60$, the most obvious change is the gradually developped wrinkles. 
Despite the big appeareance changes between adjecent age groups, the aging face should keep the identity of the input face. 
In other word, the learned transition pattern should also has the nice property of identity preserving.

\begin{figure}[t]
	\begin{center}
		\includegraphics[width=0.75\linewidth]{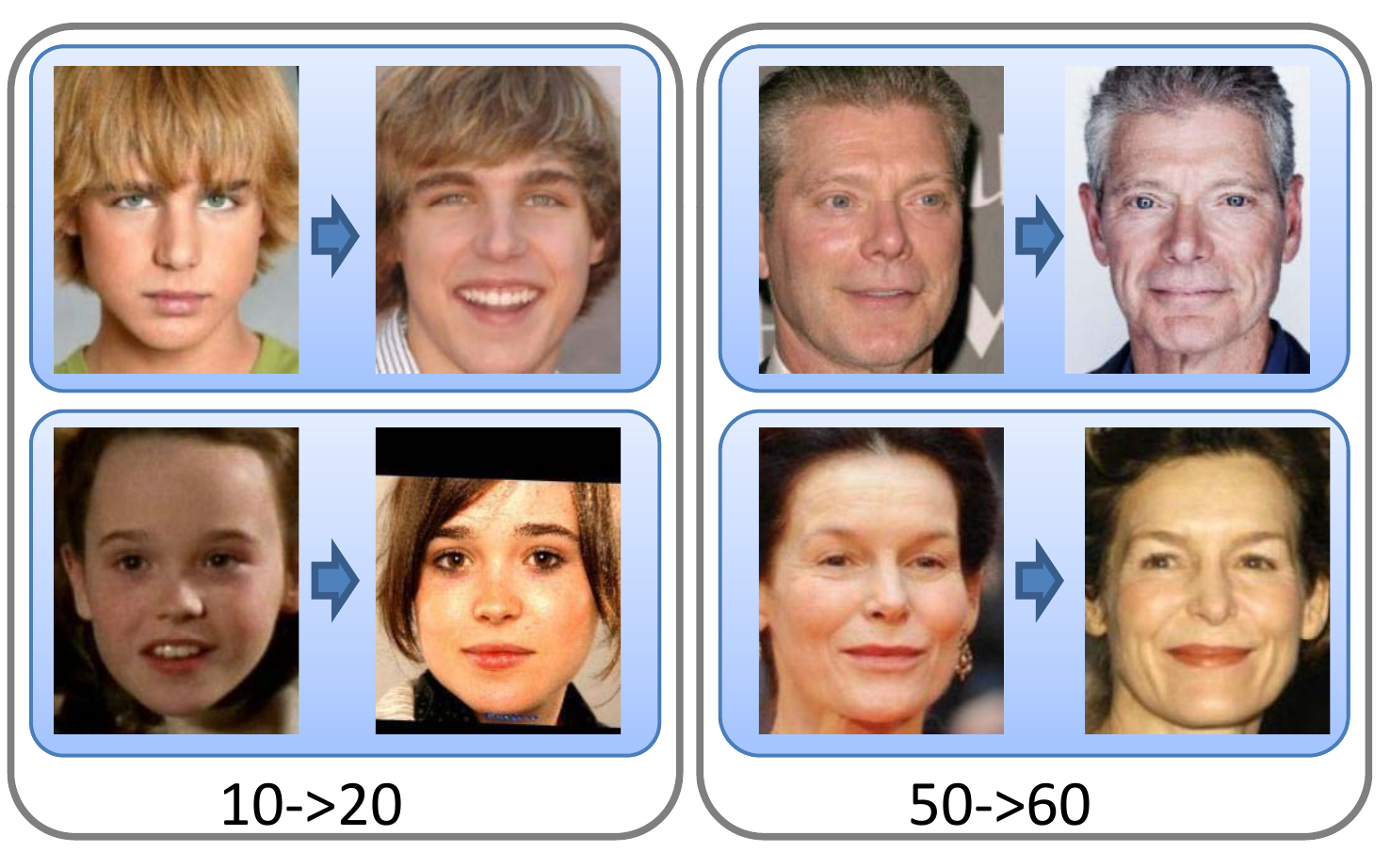}
		\vspace{-4mm}
		\caption{Illustration of the transition patterns of differnt age ranges.}
		\label{fig:transition_pattern}
		\vspace{-4mm}
	\end{center}
\end{figure}

To this end, C-GANs contains a transition pattern discriminative network as shown in Figure \ref{fig:framework}. The network ${D_t}\left( {{x_y},{x_{y + 1}},y,y + 1} \right)$ indicates  the transition pattern between ${x_y}$ aged at $y$ to the image ${x_{y + 1}}$ at the next age group $y+1$. 
For notation simplity,  the $D_t$ is denoted as  ${D_t}\left( {{x_y},{x_{y + 1}},y} \right)$. 
The networks distinguishes the real joint distribiton  ${{x_y},{x_{y + 1}},y\sim{p_{data}}({x_y},{x_{y + 1}},y)}$ from the fake one. 

\begin{figure*} [t]
	\centering
	\subfigure{
		\begin{minipage}[b]{0.8\textwidth}
			\includegraphics[width=1\textwidth]{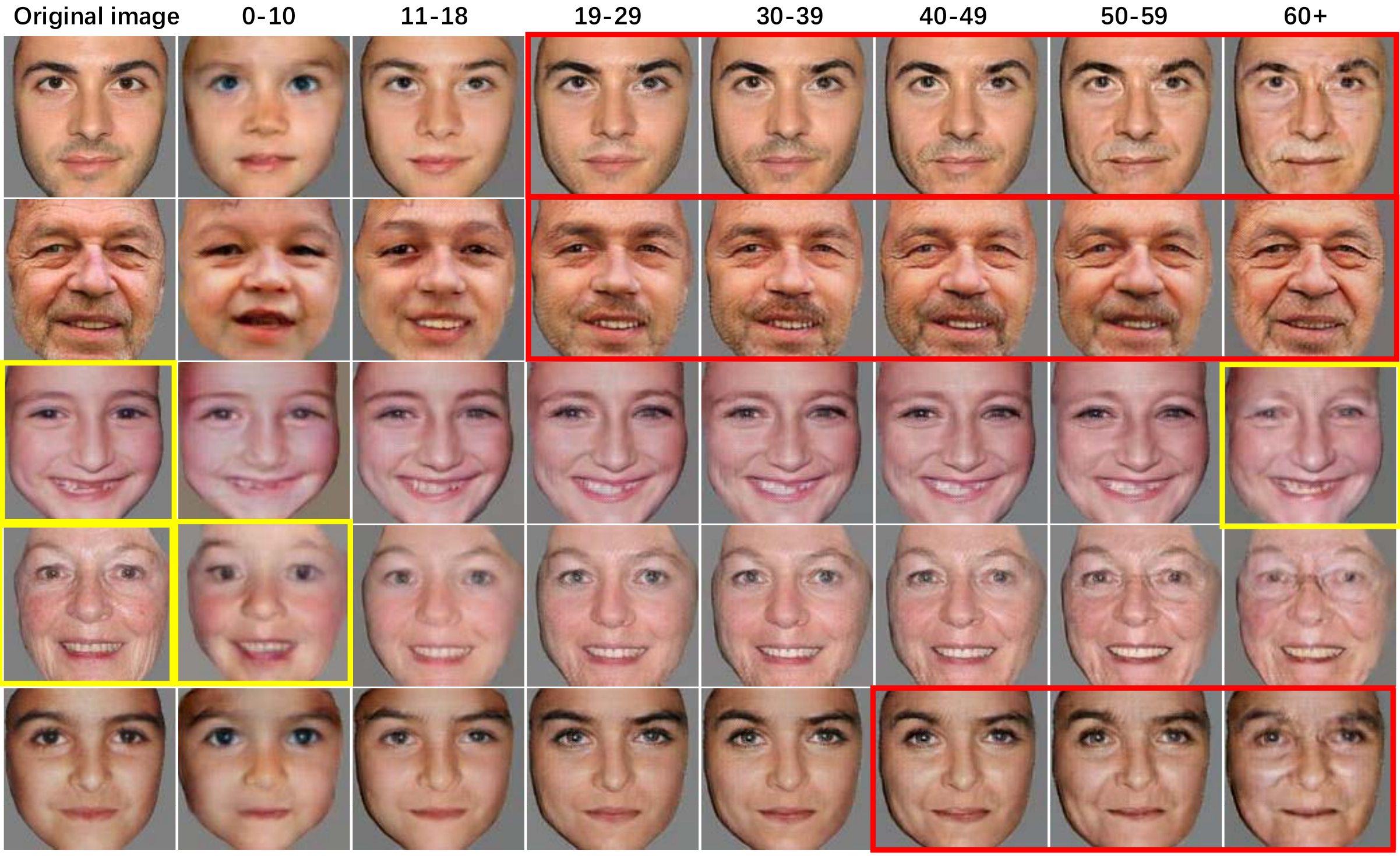}
			\vspace{-6mm}
		\end{minipage}
	}
	\subfigure{
		\begin{minipage}[b]{0.8\textwidth}
			\includegraphics[width=1\textwidth]{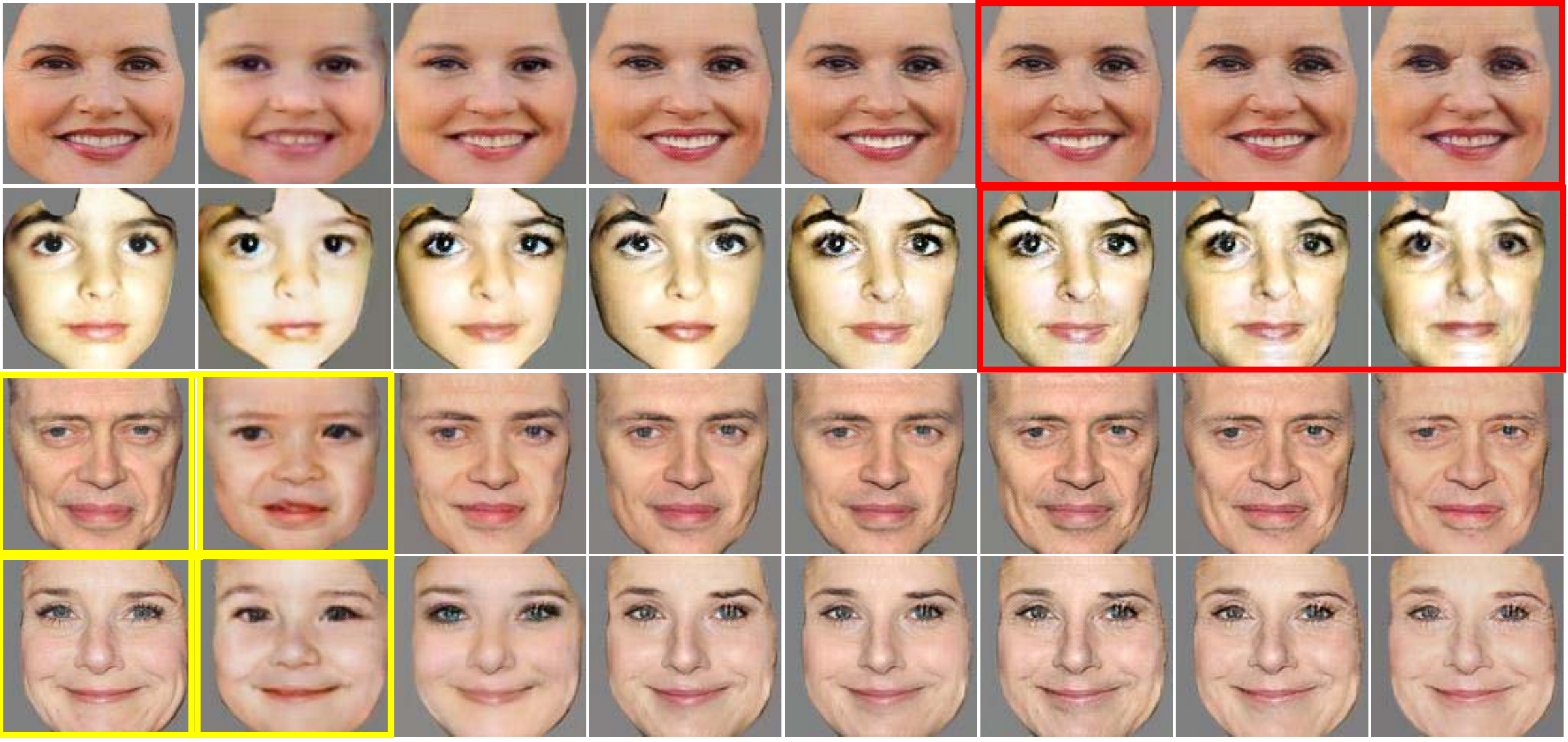}
			\vspace{-4mm}
		\end{minipage}
	}
	\vspace{-4mm}
	\caption{The input face and the generated faces for $7$  age groups. } \label{fig:face_aging_our}
	\vspace{-2mm}
\end{figure*}
The optimization objetive is:

\begin{equation}
\begin{array}{*{20}{l}}
{\begin{array}{*{20}{l}}
	{\mathop {\min }\limits_G \mathop {\max }\limits_{{D_{\rm{t}}}} E\left( {{\theta _G},{\theta _{{D_t}}}} \right){\rm{ = }}}\\
	{ = {E_{{x_y},{x_{y + 1}},y\sim{p_{data}}({x_y},{x_{y + 1}},y)}}\left[ {\log {D_t}\left( {{x_y},{x_{y + 1}},y} \right)} \right]}
	\end{array}}\\
{\begin{array}{*{20}{l}}
	{ + \frac{1}{2}{E_{{x_y},y\sim{p_{data}}({x_y},y)}}\left[ {\log \left( {1 - {D_t}\left( {{x_y},G\left( {{x_y},y + 1} \right),y} \right)} \right)} \right]}\\
	{ + \frac{1}{2}{E_{{x_y},y\sim{p_{data}}({x_y},y)}}\left[ {\log \left( {1 - {D_t}\left( {G\left( {{x_y},y - 1} \right),{x_y},y - 1} \right)} \right)} \right]}.
	\end{array}}
\end{array}
\label{eq:tp}
\end{equation}
The second term of Equation \ref{eq:tp} 
guides the transformation network to generate the fake pair  $\left\{ {{x_y},G\left( {{x_y},y + 1} \right)} \right\}$ to obey the real transition pattern distribution.  Similarily, the third term are  imposed on the tranformaion network to generate real convincing  fake pair  $\left\{ {G\left( {{x_y},y - 1} \right),{x_y}} \right\}$.

\subsection{Objective Function \& Training Strategy}  \label{sec:object_function}

Comprehensively considering the  losses of the conditional transformation network $G$, the age discriminative network $D_a$ as well as the transition pattern discriminative   network $D_t$, the overall objective function is:

\begin{equation}
\begin{array}{*{20}{l}}
{\begin{array}{*{20}{l}}
	{\mathop {\min }\limits_G \mathop {\max }\limits_{{D_{\rm{a}}}} \mathop {\max }\limits_{{D_{\rm{t}}}} E\left( {{\theta _G},{\theta _{{D_a}}},{\theta _{{D_t}}}} \right){\rm{ = }}{{\rm{E}}_a} + {E_t} + \lambda TV}\\
	{ = {E_{{x_y},y\sim{p_{data}}({x_y},y)}}\left[ {\log {D_a}\left( {{x_y},y} \right)} \right]}\\
	{ + {E_{x\sim{p_x},\tilde y\sim{p_y}}}\left[ {\log \left( {1 - {D_a}\left( {G\left( {x,\tilde y} \right),\tilde y} \right)} \right)} \right]}\\
	{{\rm{ + }}{E_{{x_y},{x_{y + 1}},y\sim{p_{data}}({x_y},{x_{y + 1}},y)}}\left[ {\log {D_t}\left( {{x_y},{x_{y + 1}},y} \right)} \right]}
	\end{array}}\\
{\begin{array}{*{20}{l}}
	{ + \frac{1}{2}{E_{{x_y},y\sim{p_{data}}({x_y},y)}}\left[ {\log \left( {1 - {D_t}\left( {{x_y},G\left( {{x_y},y + 1} \right),y} \right)} \right)} \right]}\\
	{ + \frac{1}{2}{E_{{x_y},y\sim{p_{data}}({x_y},y)}}\left[ {\log \left( {1 - {D_t}\left( {G\left( {{x_y},y - 1} \right),{x_y},y - 1} \right)} \right)} \right]}\\
	{ + \lambda \left( {TV\left( {G\left( {{x_y},y - 1} \right)} \right){\rm{ + TV}}\left( {G\left( {{x_y},y + 1} \right)} \right) + {\rm{TV}}\left( {G\left( {x,\tilde y} \right)} \right)} \right)}.
	\end{array}}
\end{array}
\end{equation}
where ${\rm{TV}}( \cdot )$ denotes the total variation which is effective in removing the ghosting artifacts. The coefficient  $\lambda$  balances the smoothness and high resolution. 
During training, $D_a$, $D_t$ and $G$ are alternatively optimized. 
More specifically, for one iteration, the $D_a$ and $G$ are updated. For the next iteration, the parameters of   $D_t$ and $G$ are refined.

\textbf{Testing Phase}  The test image ${x^t}$ and the desized label $y$  is fed into the conditional transformation network  in Section \ref{sec:transformation_network}, and the  output face $x_y^t$ is obtained. 

In the future, we would like to explore the face aging in videos by making use of  the state-of-the-art video processing  \cite{liu2016surveillance,demo2},  face parsing \cite{faceparsing}, object tracking \cite{liu2016structural,zhang2015structural} and video deblur \cite{ren2016image,ren2016single}  techniques.

\section{Experiments}

\begin{figure*} [t]
	\centering
	\subfigure{
		\begin{minipage}[b]{0.9\textwidth}
			\includegraphics[width=1\textwidth]{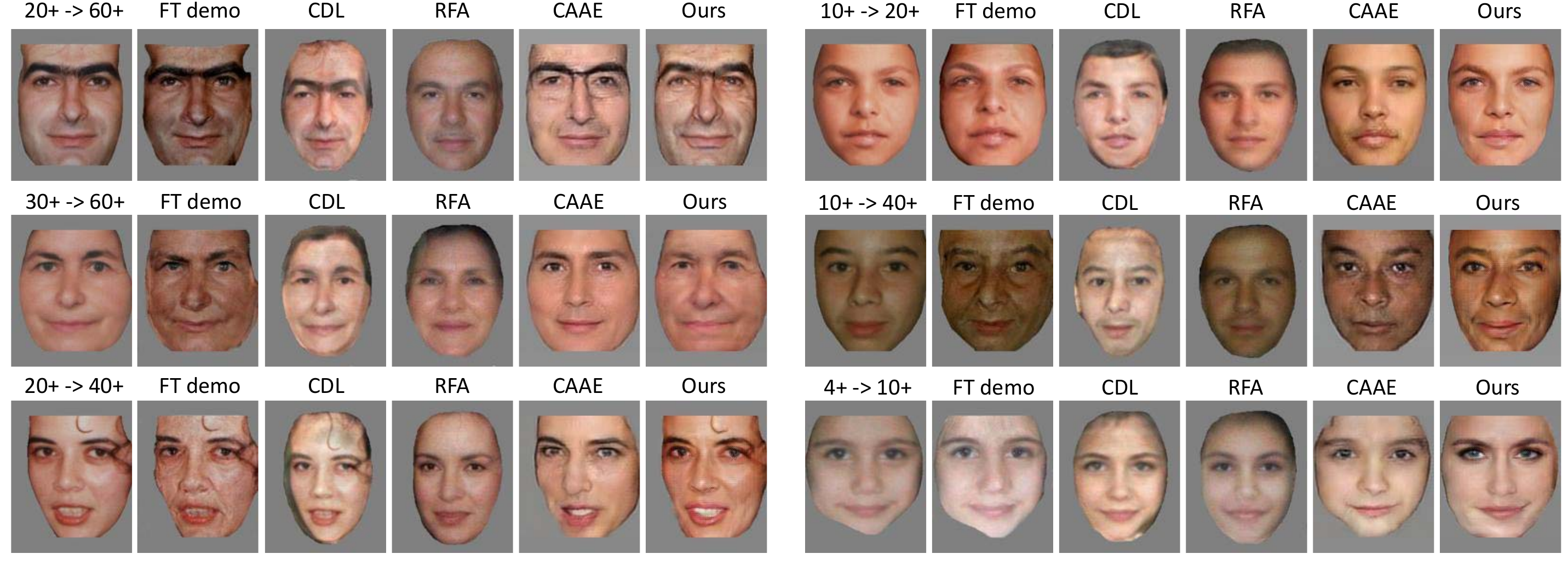}
			\vspace{-6mm}
		\end{minipage}
	}
	\subfigure{
		\begin{minipage}[b]{0.9\textwidth}
			\includegraphics[width=1\textwidth]{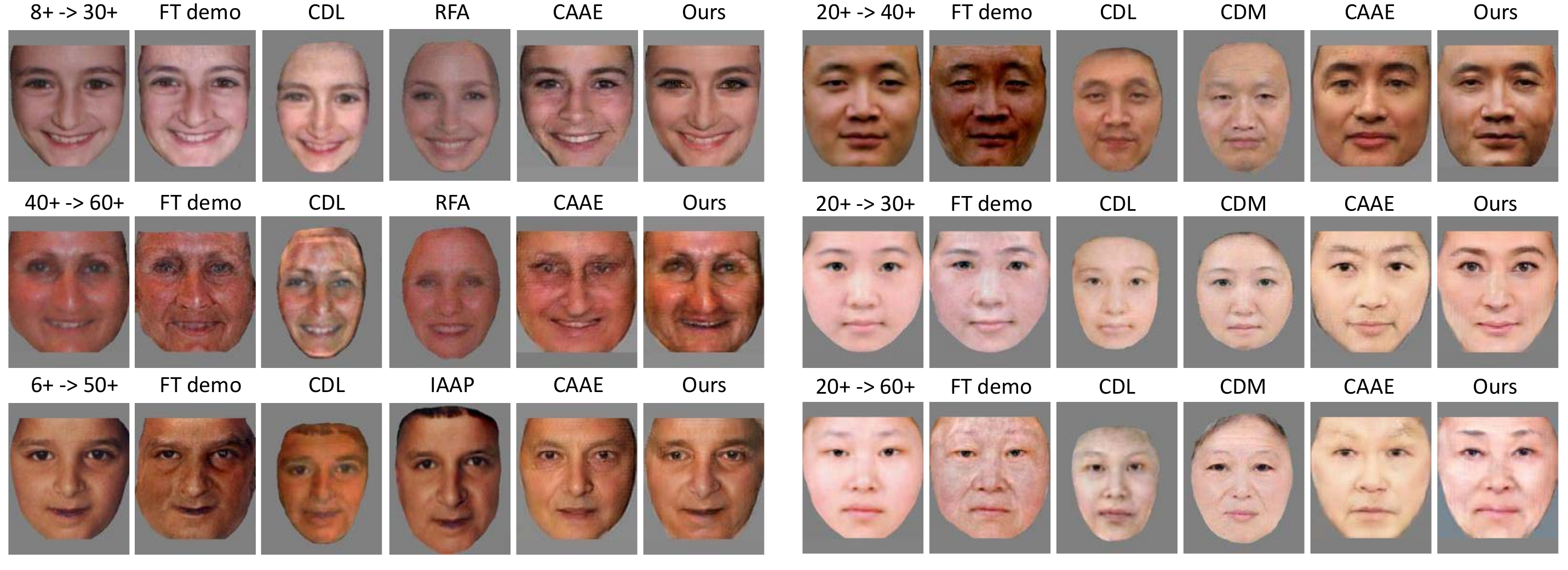}
			\vspace{-4mm}
		\end{minipage}
	}
	\caption{Comparison with the State-of-the-arts. } \label{fig:qualitive_compare_with_baseline}
\end{figure*}

\subsection{Dataset Collection}

Our C-GANs requires both sequential and non-sequential data. For sequential data, we select $575$, $649$, $1,962$, $695$ and $166$ images from CACD \cite{Chen2014Cross}, FGNET \cite{fgnet}, LFW \cite{Huang2008Labeled}, Morph \cite{Ricanek2006MORPH} and SUP \cite{Wang2016Recurrent} dataset, respectively. The whole dataset contains $4,047$ images with equal female/male and age distribution.  We generate $3,992$ positive pairs  from the sequence data for training. Note that we only use $575$ images from CACD for model training, and reserves $4,689$ images for the face verification experiment in Section  \ref{sec:verificaton}. 

For non-sequential data,  we use  the  public IMDB-Wiki dataset \cite{Rothe2015DEX}. 
We manually delete the images with heavy occlusion or  low resolution.  
As the dataset contains very few senior people, we also selected some images from the webface dataset \cite{Liu2016Deep}, 
adiencedb \cite{Eidinger2014Age}  and CACD \cite{Chen2014Cross}.  Like \cite{Antipov2017Face}, 
we divide the age into  six age categories: $0-10$, $11-18$, $19-29$, $30-39$, $40-49$, $50-59$ and $60+$ years.  The non-sequential data consists of $15,030$ face images with a uniform distribution on both gender and age.

\subsection{Implementation Details}

The C-GANs net is trained via torch based on the public available codes DCGAN \cite{Radford2015Unsupervised}. Similarly, the learning rate is set $0.0002$ and beta1 is $0.5$ with a mini-batch size $28$.  Both faces and ages are fed to the network. The generator and discriminator networks are optimized alternatively by the Adam method. Generally, the network needs $100$ epochs to generate favorable images, which may takes $12$ hour by using NVIDIA Titan X.  It takes about 0.028 sec. for testing one  image.

\subsection{Face Aging Results}

We show the face aging results in Figure \ref{fig:face_aging_our}. 
We can see that C-GANs can generate quite appealing results. 
We can draw the following observations. 
First, the faces are quite real and natural. 
Second, the generated faces can change gradually when getting older. 
For example, in the output faces of the last $5$ age groups (in red boxes) of  the first and second rows, the beards appear and become white. 
Third, C-GANs can synthesize images with large age gaps. 
For example, the input face of the third row is quite young, but the synthesized faces in the $60+$ group (in yellow box) is still quite real. 
For another example, the face in the fourth row is a senior lady. We can produce very child-looking young face (in yellow box) for the $0-10$ age group. 
Fourth, C-GANs can produce very detailed texture changes. 
For example, in the fifth, sixth and seventh rows,  the synthesized faces in the red boxes contain
frighteningly real convincing enough crow's feet,  canthus wrinkles and eye bags. 
Fifth, the  shapes of face and facial features also change during the aging/progression. 
For example, in the last two rows, when the seniors are transformed to babies, their face become smaller, and their eyes and ocular distance become larger.

\begin{figure*} [t]	\centering
	\includegraphics[width=1\textwidth]{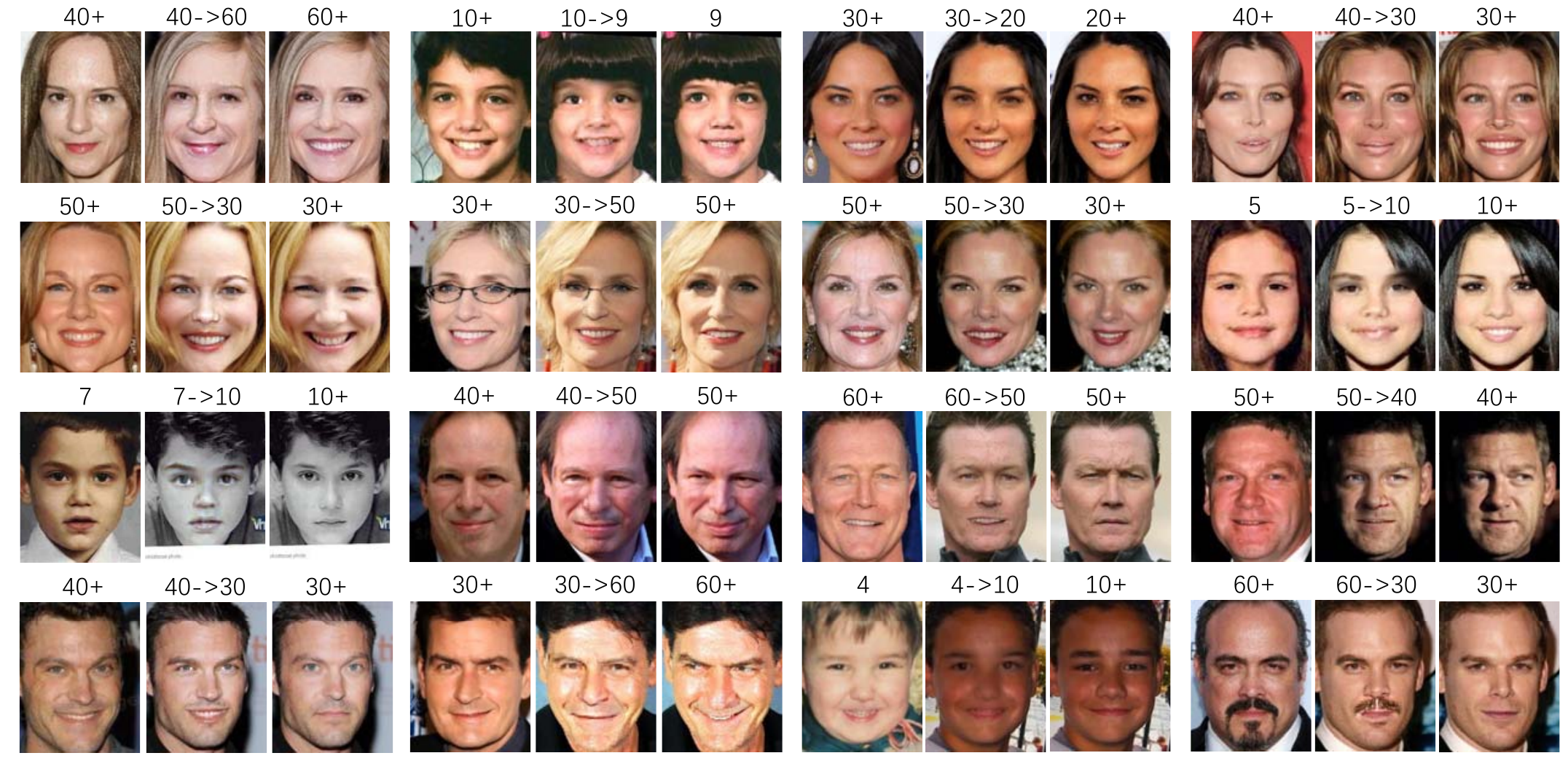}
	\vspace{-6mm}
	\caption{Comparison between our results and ground truth.} \label{fig:female_with_hair}
\end{figure*}

\subsection{Comparison with the State-of-the-arts}

\textbf{Baseline methods}: Some prior works on age progression have posted their best face aging results with inputs of different ages, 
including \cite{Suo2010A,Wang2006Age}.  
We mainly compare with $9$ baselines, including 
FT demo: an online fun demo Face Transformer, 
IAAP: state-of-the-art illumination-aware age progression
\cite{Kemelmacher2014Illumination}, RFA: recurrent face aging 
\cite{Wang2016Recurrent}, CDL: coupled dictionary learning  \cite{Shu2015Personalized},
acGAN: face aging with conditional generative adversarial networks \cite{Antipov2017Face},
CAAE:   conditional adversarial autoencoder \cite{Zhang2017Age}, 
CDM:  Compositional and Dynamic Model \cite{Suo2010A} 
and 
\cite{Sethuram2010A,Wang2006Age}. There are $246$ aging results with $72$ inputs in total. Our face aging for each input is implemented to generate the aging results with the same ages (ranges) of the posted results.

\textbf{Qualitative Evaluation:}    Figure \ref{fig:qualitive_compare_with_baseline} plots the results of the comparison. Compared with other methods, the aged face generated by our method has more realistic and noticeable transformations in appearance. For instance, the shape of facial features has changed obviously when children grow up(row 1 col 2, row 3 col 2). The aged face in the row 2 col 1 gets more wrinkles, and her eyes become smaller during the process. We can also observe that eye bags and more wrinkles appear in row 5 col 1. Meanwhile, our method preserve the identity very well, which can be observed in most images.

\begin{figure}[t]
	\begin{center}
		\includegraphics[width=0.9\linewidth]{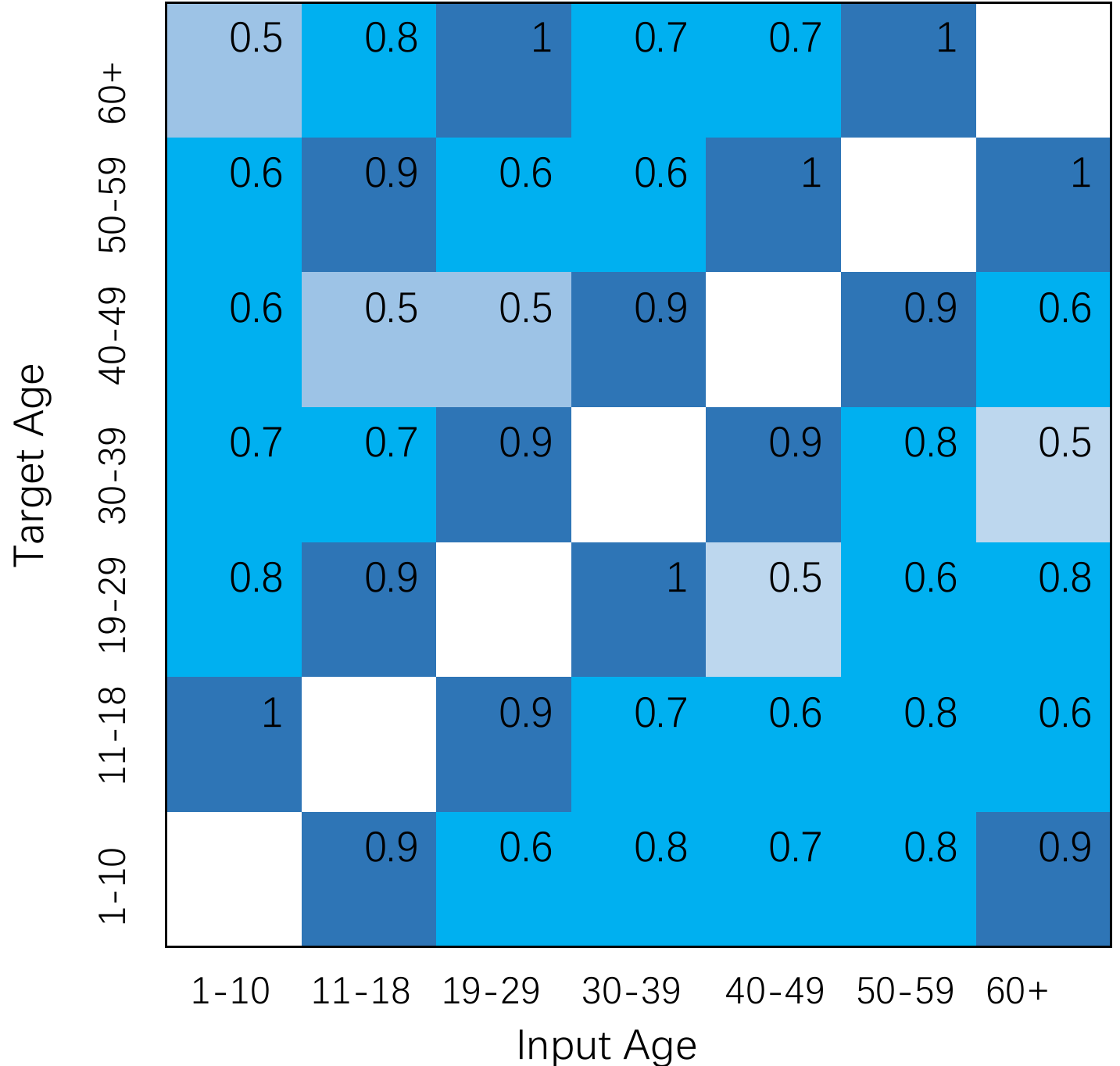}
		\caption{Comprehensive comparison to prior works.}
		\label{fig:quantitaive_compare_with_baseline}
	\end{center}
\end{figure}

\textbf{Quantitative Evaluation:} 
To quantitatively evaluate the performance of the proposed method, we designed a user study from 43 volunteers. Each volunteer is shown with three generated images each time. The candidate images are generated from given age groups and supposed to have the target ages. Among these images, there is the one generated by C-GANs and two other results generated by FT demo, CAAE or other prior methods. Every volunteer is asked to choose one of the following three options. We added 1 point if our result was chosen as the best one, 0.5 when ``all are likely" was chosen, and 0 if one result from prior work was chosen as the best. The score was normalized by number of responses per cell and shown in Figure \ref{fig:quantitaive_compare_with_baseline}. The x-axis is the input age group and the y-axis is the target age group.

From Figure \ref{fig:quantitaive_compare_with_baseline} we can see that the proposed method outperforms prior work almost all the time. Particularly,  our approach performs very good when the input and target age groups are contiguous. We believe this manly credits to the newly proposed Transition Pattern Discriminative network. We also notice that our approach is a bit poor when generating faces of 60+ from children. We think this is because of the significant changes of appearances from children to olds. We will try to improve the performance in future works.

\subsection{Comparison with ground truth}

Here we qualitatively compare  the generated faces with the ground truth. The qualitative comparisons of both female and male are shown in Figure \ref{fig:female_with_hair}, which shows appealing similarity. In each triplet, the first and third images are the ground truths with Age Group 1 and Age Group 2, while the second image is our aging result.

In this experiment, we first crop the face from images of Age Group 1. Then we proceed these images by C-GANs and get the aging faces with the same ages as Age Group 2. At last, we run graph cuts to find an optimal seam followed by poisson blending to blend the aged face into the real head photo \cite{Bitouk08}.

In Figure \ref{fig:female_with_hair}, we can observe that the generated aging faces almost have no difference from real ones. This indicates the C-GANs could correctly synthesis the real age progression.

\begin{figure}[t]
	\begin{center}
		\includegraphics[width=1\linewidth]{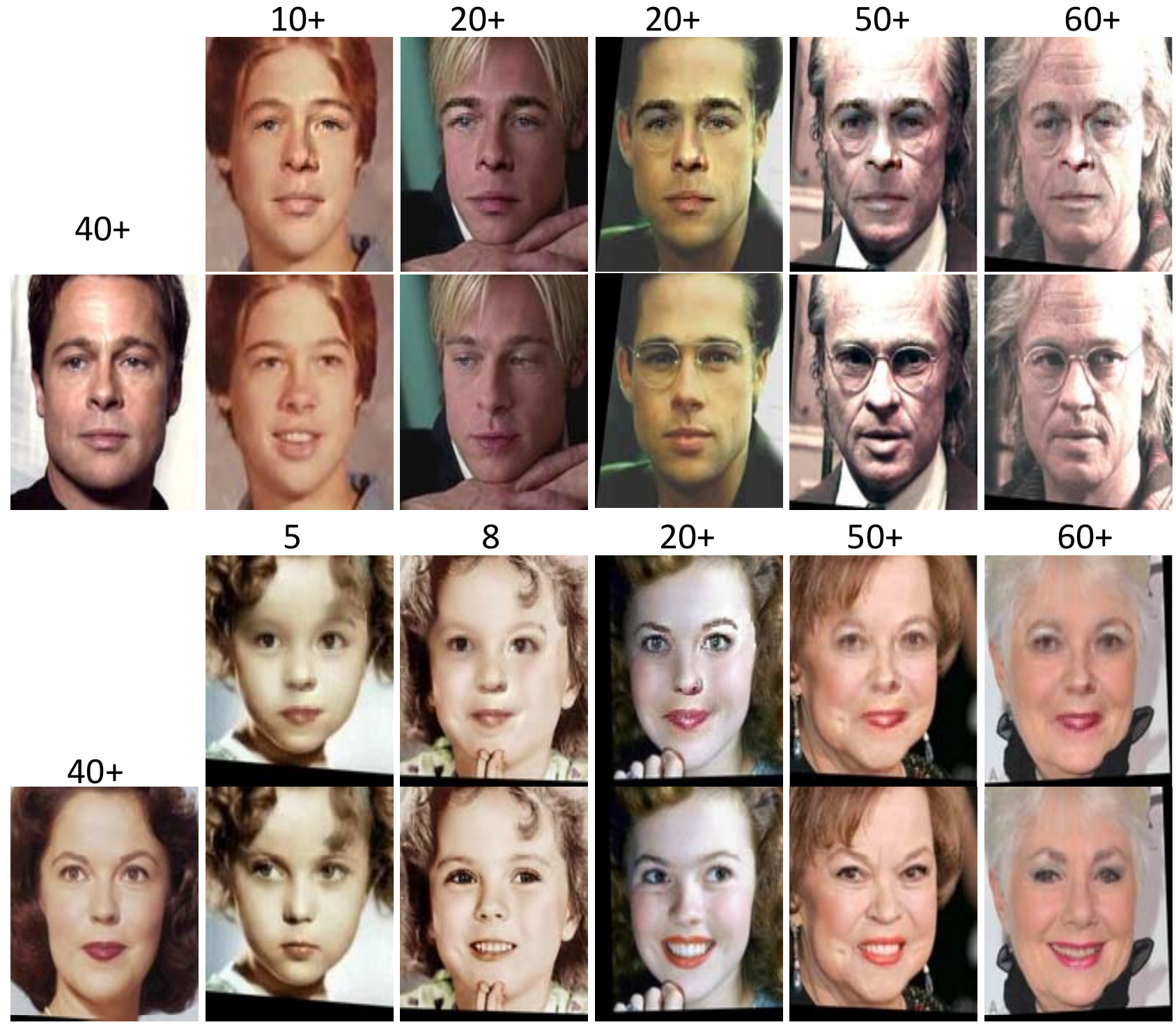}
		\caption{Movie stars. In each group, a single image on the far left is age progressed to different ages in the first row, and the real images of that age are shown in the second row.}
		\label{fig: superstar}
	\end{center}
\end{figure}

To better demonstrate the age progressing capability of C-GANs, we collect some images of two movie starts, including Brad Pitt and Shirley Temple. 
For Temple, we find several real images through her life. 
We cut out some frames from movies and take them as the older appearance of Pitt. 
The results are shown in Figure \ref{fig: superstar}. 
The age groups from $20-50$ are omitted due to the limitation of data. 
Note that our method can successfully simulate the transformation characteristics of all ages of the stars, especially for the ages of children and olds.

\subsection{Cross Age Face Verification} \label{sec:verificaton}

The proposed age progression method can also improve the performance of cross-age face verification significantly. We collected $2,000$ intra-person pairs and $2,000$ inter-person pairs with cross ages on the FGNET database, using $2,044$/$2,645$ images of males/females respectively. In both sets of pairs, the number of male pairs and female pairs are equal, and the age span of the pairs are all more than $20$ years. The set of these $4,000$ pairs is called ``Original Pairs''. In each original pair, we proceed the younger face to the aging face with the same age of the older face by C-GANs, and assume the aging face has the same age as well. We replace the younger face in original pairs by the newly generated aging face, and then construct $4,000$ new pairs, called ``Our Synthetic Pairs''. To evaluate the performance of our C-GANs, we also generated the ``CAAE Synthetic Pairs'' by the state-of-the-art age progression method \cite{Zhang2017Age}. The state-of-the-art center Loss based face verification \cite{Wen2016A} is used for testing on the above three sets of pairs.

\begin{figure}[t]
	\begin{center}
		\includegraphics[width=1\linewidth]{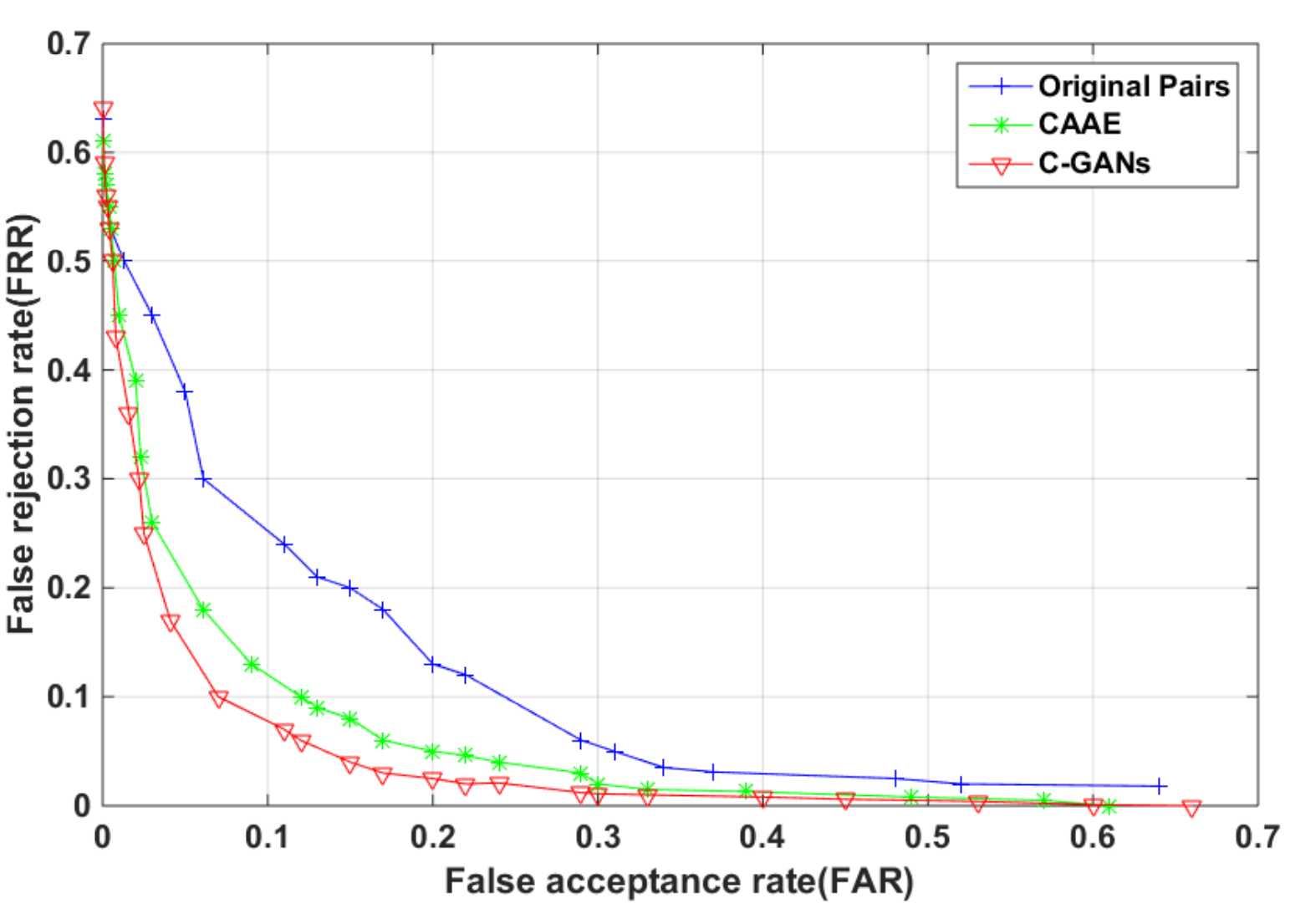}
		\caption{Face verification results. }
		\label{fig:face_verification}
	\end{center}
\end{figure}

The FAR-FRR (false acceptance rate-false rejection rate) curves are illustrated in Figure \ref{fig:face_verification}. The EER (the equal error rates) on C-GANs Synthetic Pairs, CAAE Synthetic Pairs, and Original Pairs are $8.72\%$, $11.05\%$, and $17.41\%$ respectively.  We can see that the face verification on C-GANs Synthetic pairs achieves better ERR than on other pairs. This implies the aging faces generated by C-GANs can effectively alleviate the face verification errors cost by age gaps.

\section{Conclusion and Future Works}
In this paper, we propose a contextual generative adversarial nets to tackle the  face aging  problem. 
Different from existing generative adversarial nets based methods, we explicitly model the transition patterns  between adjacent age groups during the training procedure.  From baby to teenagers period, the transition patterns is shown in the way that the face becomes bigger, while from  the ages of $30$ to the age of $50$, the transition patterns include the  gradually developed wrinkle. To this end, the C-GANs consists of two discriminative networks, i.e., an age discriminative network and a transition pattern discriminative network. They are collaboratively contribute to the appealing results.  

Currently, our model is based on DCGAN \cite{Radford2015Unsupervised}.  
In future, we plan to employ other GANs to improve the performance, such as  Wasserstein GAN \cite{Arjovsky2017Wasserstein}, LS GAN \cite{Guo2017Loss}, EB GANs \cite{Junbo2017Energy}.

\section{Acknowledgment}
This work was supported by  National Natural Science Foundation of China (No.61572493, Grant U1536203) and   Natural Science Foundation of Jiangsu Province (Grant No. BK20170856) 
	 
\bibliographystyle{ACM-Reference-Format}
\balance
\bibliography{clothes} 
	
\end{document}